\documentclass[11pt]{article}
\usepackage[preprint]{acl}
\usepackage{times}
\usepackage{latexsym}
\usepackage[T1]{fontenc}
\usepackage[utf8]{inputenc}
\usepackage{microtype}
\usepackage{graphicx}
\usepackage{booktabs}
\usepackage{url}
\usepackage{xcolor}
\usepackage{amsmath}
\usepackage{amssymb}
\usepackage{tikz}
\usetikzlibrary{arrows.meta,positioning}

\title{Competence-Preserving Resume Perturbations Expose Presentation Sensitivity in LLM Screening}
\author{
  Qiangju Chen$^1$,Yang Xiao$^2$\\
  $^1$Macquarie University  \\
  $^2$The University of Melbourne \\
\texttt{qiangju.chen@students.mq.edu.au}
}

\begin{document}
\maketitle
\begin{abstract}
Resume screeners must infer job-relevant competence from resumes whose
presentation can vary substantially in wording, structure, stylistic polish,
and document extraction quality. Ideally, such surface variation should not
change decisions when the underlying qualification evidence is unchanged.
We introduce a controlled audit of this property, constructing
occupation-grounded candidate profiles at controlled competence levels and
rendering each profile into multiple resume presentations. A deterministic
validation gate excludes variants that alter the underlying evidence before
scoring. Across six open instruction-tuned LLM conditions, we find a clear
disconnect between screening validity and presentation stability.
Llama-3.1-8B with its native chat template achieves the strongest validity
($0.781$) yet reverses $29.6\%$ of matched pairwise decisions under
competence-preserving presentation changes; Mistral-7B-v0.3 reaches validity
$0.644$ with a $41.4\%$ flip rate. Native chat formatting improves validity
for several chat-tuned models but does not remove this instability. These
results show that resume-screening evaluations should assess not only whether
a system identifies stronger candidates, but also whether those decisions
remain stable when the same competence evidence is presented differently.
\end{abstract}

\section{Introduction}


Automated resume screening requires systems to infer job-relevant
competence from highly variable natural-language documents. The same
qualification evidence can appear in concise bullet points, paragraph-style
descriptions, AI-polished prose, or text affected by document extraction and
layout artifacts. These differences are often orthogonal to whether a
candidate is actually qualified, yet they change the surface form presented
to the screening model. A central challenge is therefore to distinguish
variation in candidate competence from variation in how that competence is
expressed.

Algorithmic hiring has long raised questions about the validity,
fairness, and reliability of automated decision systems
\citep{raghavan2020mitigating,kochling2020discriminated,
dearteaga2019bias}. More recent work on LLM-based hiring has examined
screening validity \citep{castleman2026validity}, demographic bias
\citep{gao2026fairly,iso2025evaluating}, self-preference for
AI-written resumes \citep{xu2025selfpref}, and the influence of AI
recommendations on human screeners \citep{wilson2026fastslow}.
These studies address important questions about whether screening
systems make useful or fair decisions, but leave open a complementary
robustness question: \emph{when the underlying competence evidence is
unchanged, does the screening decision remain stable under changes in
presentation?} This question is practically relevant because applicants
differ in writing assistance, resume templates, formatting conventions,
and document-conversion pipelines. Presentation variation is therefore
not merely an artificial perturbation, but a natural source of
heterogeneity in the inputs that screening systems receive.

We study this problem through a controlled O*NET-backed audit that
separates candidate competence from resume presentation. Candidate
profiles are constructed from occupation-grounded rubrics at controlled
competence levels and then rendered into multiple resume forms. A
deterministic validation gate removes variants that alter the underlying
evidence before scoring. This design allows us to evaluate two properties
on the same candidate set: whether a screener recovers known-superiority
ordering (\emph{validity}) and whether those decisions survive
competence-preserving presentation changes (\emph{presentation
invariance}). We find that the two properties can diverge sharply:
Llama-3.1-chat achieves validity $0.781$ yet reverses $29.6\%$ of
matched decisions, while Mistral reaches validity $0.644$ with a
$41.4\%$ flip rate. Native chat formatting improves validity for several
chat-tuned models but does not eliminate this sensitivity, consistent
with broader evidence that LLM behavior can depend strongly on prompt
format and other semantically incidental prompt choices
\citep{sclar2024quantifying,chatterjee2024posix}.

\section{Benchmark Framework}

\begin{figure}[t]
\centering

\includegraphics[width=\columnwidth]{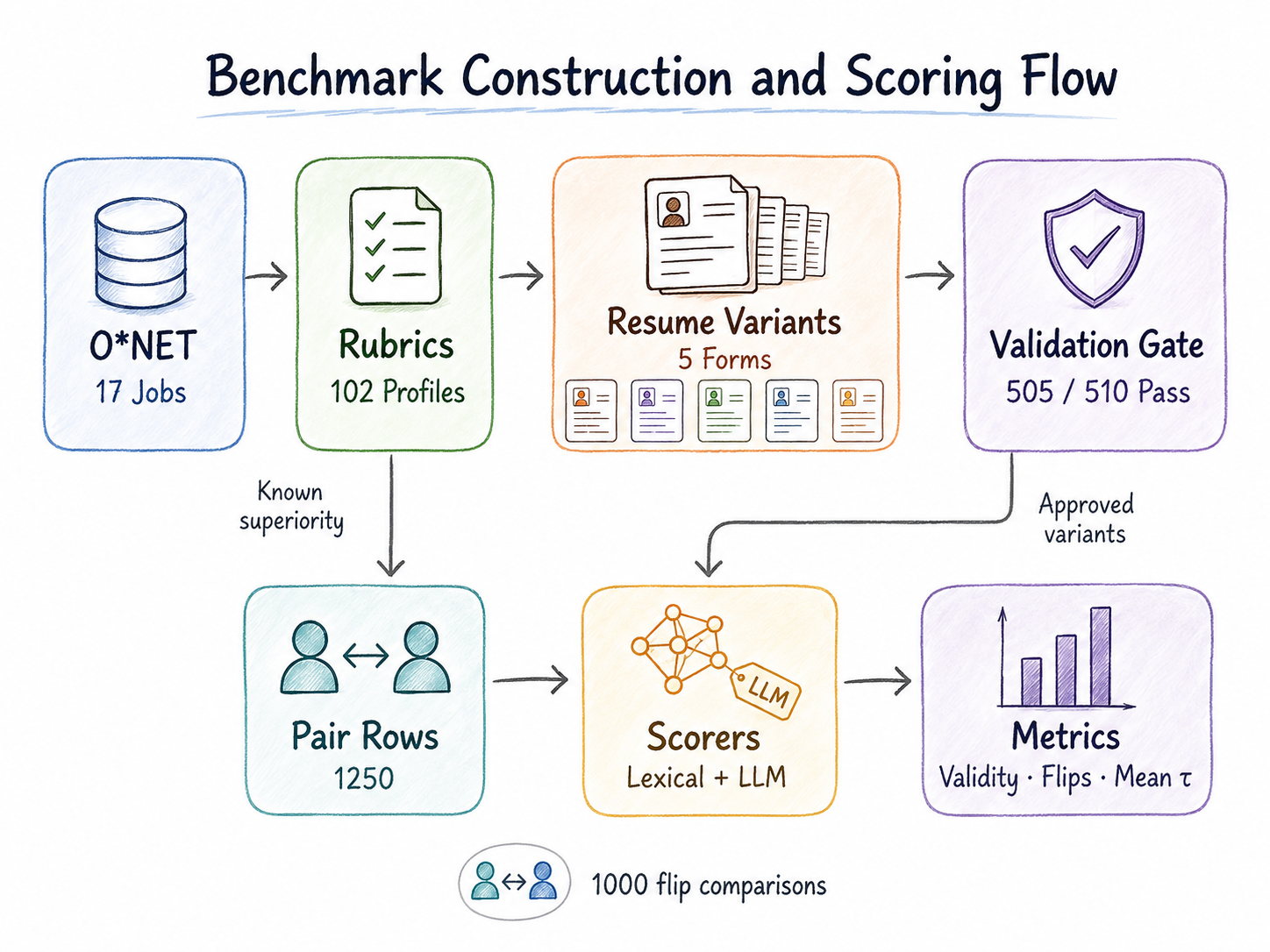}
\caption{Benchmark construction and scoring flow. }
\label{fig:benchmark}
\vspace{3mm}
\end{figure}

Our benchmark is designed to isolate a single question: \textbf{does a resume screener preserve its decision when competence is fixed but presentation changes?} To make this measurable, we separate benchmark construction into three stages as Figure \ref{fig:benchmark}. We first instantiate occupation-grounded candidate profiles with controlled competence differences, then render each profile into multiple presentation forms, and finally admit only variants that pass deterministic fact-preservation checks. Scorers operate on the validated resume text rather than hidden competence labels. This design allows known-superiority validity and presentation invariance to be evaluated on the same underlying candidate set.

\subsection{O*NET-Grounded Candidate Profiles}

We construct the benchmark from the O*NET 30.3~\citep{onet2026} database, selecting 17
occupations spanning technical, administrative, customer-facing, and
health-related roles. For each occupation, we derive a screening rubric
from its title, required skills, knowledge elements, and representative
core tasks. We then instantiate six synthetic candidate profiles for a total of 102 profiles, per
occupation, with two qualified, two borderline, and two under-qualified. Profiles differ in rubric-aligned skill evidence,
years of experience, and achievement strength, providing controlled
competence differences from which known-superiority comparisons can be
constructed.

This construction separates candidate competence from its eventual textual
realization. Competence is specified at the profile level before any resume
is rendered, so multiple presentation variants can later be generated from
the same underlying candidate record. Known-superiority pairs support the
validity evaluation, while same-tier pairs provide controlled
equal-competence comparisons. This design allows presentation changes to be
studied without redefining the underlying candidate qualifications.

\subsection{Competence-Preserving Resume Perturbations}

Each candidate profile is rendered into five resume forms: an original
bullet-style resume and four controlled presentation perturbations.
\textbf{Verbosity} adds redundant summary language around the same facts;
\textbf{structure} reorganizes bullet-style evidence into paragraph-like
prose; \textbf{AI polish} rewrites tone and fluency without introducing new
skills; and \textbf{layout} simulates extraction artifacts and local ordering
noise. Together, these axes vary length, discourse organization, stylistic
polish, and document-extraction quality while keeping the intended
job-relevant evidence fixed. Across 102 candidate profiles, this process
produces 510 resume variants.

Because an intended rewrite is not necessarily competence-preserving, every
generated variant passes through a deterministic validation gate before
scoring. A variant is accepted only if required fact fields remain present,
no unsupported rubric skill is introduced, no hidden competence-tier label
is exposed, and perturbation generation is independent of the downstream
scorer. Of the 510 variants, 505 pass validation. For same-tier base pairs,
we additionally require exact agreement in the benchmark evidence vector
(required-skill hits, task hits, and years of experience). The 255 base
candidate comparisons then expand across presentation axes into 1,250
validation-approved pair rows. For the invariance audit, each non-original
axis is matched to its corresponding original-axis decision, yielding 1,000
flip comparisons.

\subsection{Independent Screening Protocol}

Each resume is scored independently against the rubric of its corresponding
occupation. Given a resume $r_i$ and occupation rubric $q$, a screening
function produces a scalar score.

\begin{equation}
     s_i = f(r_i, q).
\end{equation}

For two candidates $i$ and $j$, the pairwise decision is computed only after both resumes have been scored:

\begin{equation}
        D(i,j) = \operatorname{sign}(s_i - s_j).
\end{equation}

The scorer therefore never receives two resumes in the same comparison
prompt. This avoids pair-order and comparative prompt-position effects in
the known-superiority evaluation. It is also important for the invariance
audit: an original-to-perturbed decision change reflects movement in
independently assigned resume scores rather than a change in the surrounding
pairwise prompt context.

\subsection{Evaluation Metrics}

We evaluate screening systems along two primary dimensions:
\textbf{validity} and \textbf{presentation invariance}.

\paragraph{Known-superiority validity.}
For a known-superiority pair $(i,j)$, where candidate $i$ is constructed to
be stronger than candidate $j$, validity measures whether the scorer assigns
the stronger candidate the higher score:
\[
    \mathrm{Validity}
    =
    \frac{1}{N}
    \sum_{(i,j)}
    \mathbb{I}[s_i > s_j].
\]
Validity therefore measures recovery of the benchmark's controlled
competence ordering.

\paragraph{Presentation flip rate.}
For each non-original presentation axis $a$, we compare its pairwise
decision $D_a(i,j)$ with the decision obtained from the corresponding
original resumes, $D_0(i,j)$:

\begin{equation}
        \mathrm{Flip}
    =
    \frac{1}{M}
    \sum_{(i,j,a)}
    \mathbb{I}[D_a(i,j) \neq D_0(i,j)].
\end{equation}

A flip indicates that the pairwise decision changes when presentation
changes while the underlying competence record is held fixed. Flip rate
measures \emph{decision instability}, not error: a flip may either introduce
an incorrect decision or correct an originally incorrect one. We therefore
treat validity and flip rate as complementary properties.

\paragraph{Complementary stability metrics.}
For same-tier pairs, we report the fraction assigned equal scores as the
equal-tier tie rate. We also compute Kendall's $\tau$ between original-axis
and perturbed-axis rankings within each occupation and average it across
occupations. Flip rate captures matched pairwise decision changes, whereas
Kendall's $\tau$ provides an occupation-level view of ranking stability.


\section{Experimental Setup}

\paragraph{Screening systems.}
We evaluate lexical baselines and direct LLM scorers under the
validation-gated benchmark. The lexical baselines are BM25
\citep{robertson2009bm25} and TF--IDF
\citep{salton1988term}. The direct LLM scorers span five model
families: Qwen2.5-7B \citep{qwen2025}, Llama-3.1-8B and
Llama-3.2-3B \citep{llama2024}, Phi-3.5-mini
\citep{phi2024}, Mistral-7B-v0.3
\citep{mistral2023}, and Gemma-2-2B
\citep{gemma2024}. 

We additionally include a deterministic phrase-preserving
control that matches rubric skill and task phrases together
with years of experience and aggregates the resulting evidence
deterministically.

\paragraph{Prompt conditions.}
We use raw prompts where available and native chat templates
for Llama-3.1, Llama-3.2, and Gemma as a prompt-format
ablation. The main table reports the stronger native-template
condition for Llama-3.2 and Gemma.

\section{Results}

\begin{table*}[t]
\centering
\small
\caption{Validity and presentation stability across screening systems.
Higher validity indicates better recovery of known-superiority ordering,
while lower flip rate indicates greater stability under
competence-preserving presentation changes.}
\vspace{3mm}
\begin{tabular}{lrrrr}
\toprule
System & Validity & Cluster 95\% CI & Flip & Mean $\tau$ \\
\midrule
BM25 lexical & 0.488 & [0.367, 0.603] & 0.040 & 0.992 \\
TF-IDF lexical & 0.476 & [0.375, 0.584] & 0.052 & 0.979 \\
Qwen2.5-7B raw & 0.555 & [0.470, 0.630] & 0.377 & 0.682 \\
Llama-3.1-8B raw & 0.622 & [0.513, 0.712] & 0.357 & 0.708 \\
Llama-3.1-8B chat & \textbf{0.781} & [0.746, 0.814] & 0.296 & 0.755 \\
Phi-3.5-mini raw & 0.611 & [0.563, 0.663] & 0.311 & 0.732 \\
Mistral-7B-v0.3 raw & 0.644 & [0.593, 0.691] & 0.414 & 0.630 \\
Llama-3.2-3B chat & 0.551 & [0.488, 0.614] & 0.285 & 0.762 \\
Gemma-2-2B chat & 0.576 & [0.506, 0.644] & 0.456 & 0.590 \\
\bottomrule
\end{tabular}

\label{tab:main}
\end{table*}


\subsection{Validity and Presentation Invariance}

Table~\ref{tab:main} shows that validity and presentation
invariance capture distinct properties of a screening system.
The strongest direct LLM condition is Llama-3.1-8B with its
native chat template, which achieves validity 0.781. Under paired occupation-cluster bootstrap,
it exceeds BM25 by 0.293 and TF--IDF by 0.305. Mistral-7B-v0.3 also clears both lexical baselines, reaching
validity 0.644. Phi-3.5-mini clears TF--IDF but not BM25, while the remaining direct LLM conditions provide
directional rather than confirmatory validity evidence.

Stronger validity, however, does not imply presentation
invariance. Llama-3.1-chat reverses $29.6\%$ of matched
pairwise decisions under competence-preserving presentation
changes, while Mistral flips $41.4\%$. The same pattern is
visible across the broader direct LLM set, with flip rates
ranging from 0.285 to 0.456. Thus, even systems that
recover known-superiority ordering relatively well can remain
substantially sensitive to how the same competence evidence is
presented.

This separation motivates treating validity and presentation
invariance as complementary evaluation dimensions. Validity
measures whether a screener recovers the intended competence
ordering, whereas invariance measures whether that decision
survives irrelevant surface variation. High invariance alone
is therefore not sufficient: a system may preserve consistently
poor decisions. The desirable regime is one in which screening
decisions are both valid and stable.

BM25 and TF--IDF show much lower flip rates (0.040 and
0.052), but these values should not be interpreted as
evidence that lexical matching is generally a better resume
screener. The benchmark deliberately preserves many rubric
phrases across presentation variants, allowing lexical systems
to retain the same pairwise sign under this controlled
phrase-preserving condition. They therefore serve as stability
references rather than deployment-quality screening methods.
The deterministic structured control is narrower still: its
validity 1.000 and flip rate 0.000 arise by construction
and serve only as a benchmark sanity check.

\subsection{Prompt-Format Ablation}

A potential confound is that chat-tuned models may be
miscalibrated under raw completion prompts. Table~\ref{tab:chat}
tests this explanation using native chat templates for
Llama-3.1, Llama-3.2, and Gemma. Chat formatting substantially
improves validity: Llama-3.1 rises from 0.622 to 0.781,
Llama-3.2 from 0.321 to 0.551, and Gemma from 0.287 to
0.576. Weak raw-prompt validity should therefore not be
interpreted as an intrinsic inability of these chat-tuned
models to perform the task.

The presentation-sensitivity result nevertheless survives this
correction. Native-template flip rates remain 0.296 for
Llama-3.1, 0.285 for Llama-3.2, and 0.456 for Gemma.
Prompt-format mismatch can therefore explain part of the
validity degradation under raw prompting, but it is not a
sufficient explanation for presentation instability.


\begin{table}[t]
\centering
\scriptsize
\vspace{3mm}
\caption{Native chat-template ablation. Raw completion prompts can understate validity for chat-tuned models, but presentation sensitivity remains high after the correction.}
\resizebox{\columnwidth}{!}{%
\begin{tabular}{lrrrr}
\toprule
Model & Raw val. & Chat val. & Raw flip & Chat flip \\
\midrule
Llama-3.1-8B & 0.622 & 0.781 & 0.357 & 0.296 \\
Llama-3.2-3B & 0.321 & 0.551 & 0.363 & 0.285 \\
Gemma-2-2B & 0.287 & 0.576 & 0.363 & 0.456 \\
\bottomrule
\end{tabular}
\vspace{3mm}
}

\label{tab:chat}
\end{table}

\section{Conclusion}
We introduced a controlled audit for studying how resume-screening systems
respond to presentation variation when job-relevant competence is held fixed.
By separating candidate competence from resume realization, the benchmark
makes it possible to evaluate screening validity and presentation stability
on the same underlying candidates. Our results show that these properties
can diverge substantially: the strongest-validity LLM conditions still
reverse a large fraction of decisions under competence-preserving rewrites,
and native chat formatting does not eliminate the effect. These findings
highlight presentation stability as an important dimension of resume-screening
evaluation. Future audits should therefore examine not only whether a system
ranks stronger candidates correctly, but also whether those decisions remain
consistent across equivalent forms of the same evidence.

\clearpage

\section{Limitations}

Our controlled construction trades breadth for attribution: the benchmark
covers 17 occupations and 102 synthetic O*NET-grounded candidates, allowing
presentation effects to be isolated while competence evidence is held fixed.
Extending the audit to more occupations, human-authored resumes, and
production resume-processing pipelines would test how well these findings
generalize beyond the controlled setting. The occupation-cluster analysis is
also based on 17 clusters, so its bootstrap intervals are best interpreted as
pilot-scale evidence rather than precise population estimates. Finally, the
evaluated systems are open-weight instruction models; broader audits could
include proprietary screening models and end-to-end applicant-tracking
systems.

\bibliography{references}

\clearpage
\appendix

\section{Implementation Details}

Phi-3.5-mini and Mistral-7B-v0.3 checkpoints are obtained
from Hugging Face. Llama, Phi, Mistral, and the evaluated
chat-template conditions are executed through vLLM on a
single NVIDIA L40S GPU with 48\,GB of memory. Each resume is
scored independently against its corresponding occupation
rubric; no model receives both resumes from a comparison pair
in the same prompt. Pairwise decisions are formed only after
the individual scalar scores have been produced, avoiding
pair-order and comparative prompt-position effects.

For uncertainty estimation, we report both row-level bootstrap
intervals and occupation-cluster percentile bootstrap intervals
using 10,000 resamples. The cluster bootstrap resamples the
17 occupations and serves as a robustness check against
row-level pseudo-replication. Because the benchmark contains
only 17 occupation clusters, we interpret these intervals as
exploratory small-cluster evidence rather than precise nominal
population coverage.
\section{Axis-Level Presentation Sensitivity}

\begin{figure}[t]
\centering
\includegraphics[width=\columnwidth]{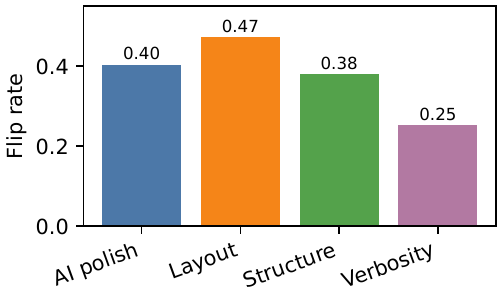}
\caption{Axis-level flip-rate diagnostic for a directional raw-model slice. AI-polished prose and layout artifacts are among the most unstable axes in this slice.}
\label{fig:axis}
\end{figure}

As a directional failure-analysis slice, Figure~\ref{fig:axis}
decomposes Qwen2.5-7B raw flips by presentation axis. Layout
artifacts produce the highest flip rate (0.472), followed by
AI-polished prose (0.404), structural reorganization
(0.380), and verbosity (0.252). Because Qwen's validity advantage over the lexical baselines
does not clear zero under paired occupation-cluster bootstrap,
we treat this decomposition as an illustrative diagnostic rather
than evidence that these axes have a universal ordering of
difficulty.

\end{document}